\documentclass[11pt]{article}
\newtheorem{problem}{Problem}
\usepackage{algorithm}
\usepackage{algorithmic}
\usepackage[font=small,labelfont=bf]{caption}
\usepackage{wrapfig}

\usepackage{cite}
\usepackage{amsmath,amssymb,amsfonts}
\usepackage{algorithmic}
\usepackage{graphicx}
\usepackage{textcomp}
\usepackage{xcolor}
\usepackage[margin=1in]{geometry}
\usepackage{algorithm}
\usepackage{algorithmic}
\usepackage[font=small,labelfont=bf]{caption}
\usepackage{wrapfig}
\usepackage{authblk}

\begin{document}


\title{Targeted Adversarial Attacks against \\ Neural Network Trajectory Predictors}
\author{Kaiyuan Tan}
\author{Jun Wang}
\author{Yiannis Kantaros\thanks{The authors are with the Department of Electrical and Systems Engineering, Washington University in St. Louis, Saint Louis, MO 63108, USA. Email:{\tt\small \{kaiyuan.t, junw, ioannisk\}@wustl.edu}}
}
\affil{Department of Electrical and Systems Engineering, Washington University in St. Louis}

\date{}
\maketitle 



\begin{abstract}%

Trajectory prediction is an integral component of modern autonomous systems as it allows for envisioning future intentions of nearby moving agents. Due to the lack of other agents' dynamics and control policies, deep neural network (DNN) models are often employed for trajectory forecasting tasks. Although there exists an extensive literature on improving the accuracy of these models, there is a very limited number of works studying their robustness against adversarially crafted input trajectories. To bridge this gap, in this paper, we propose a targeted adversarial attack against DNN models for trajectory forecasting tasks. We call the proposed attack TA4TP for Targeted adversarial Attack for Trajectory Prediction. Our approach generates adversarial input trajectories that are capable of fooling DNN models into predicting user-specified target/desired trajectories. Our attack relies on solving a nonlinear constrained optimization problem where the objective function captures the deviation of the  predicted trajectory from a target one while the constraints model physical requirements that the adversarial input should satisfy. The latter ensures that the inputs look natural and they are safe to execute (e.g., they are close to nominal inputs and away from obstacles). We demonstrate the effectiveness of TA4TP on two state-of-the-art DNN models and two datasets. To the best of our knowledge, we propose the first targeted adversarial attack against DNN models used for trajectory forecasting. 
\end{abstract}

\section{Introduction} 
\begin{figure}[t]
  \centering
    \includegraphics[width=1\linewidth]{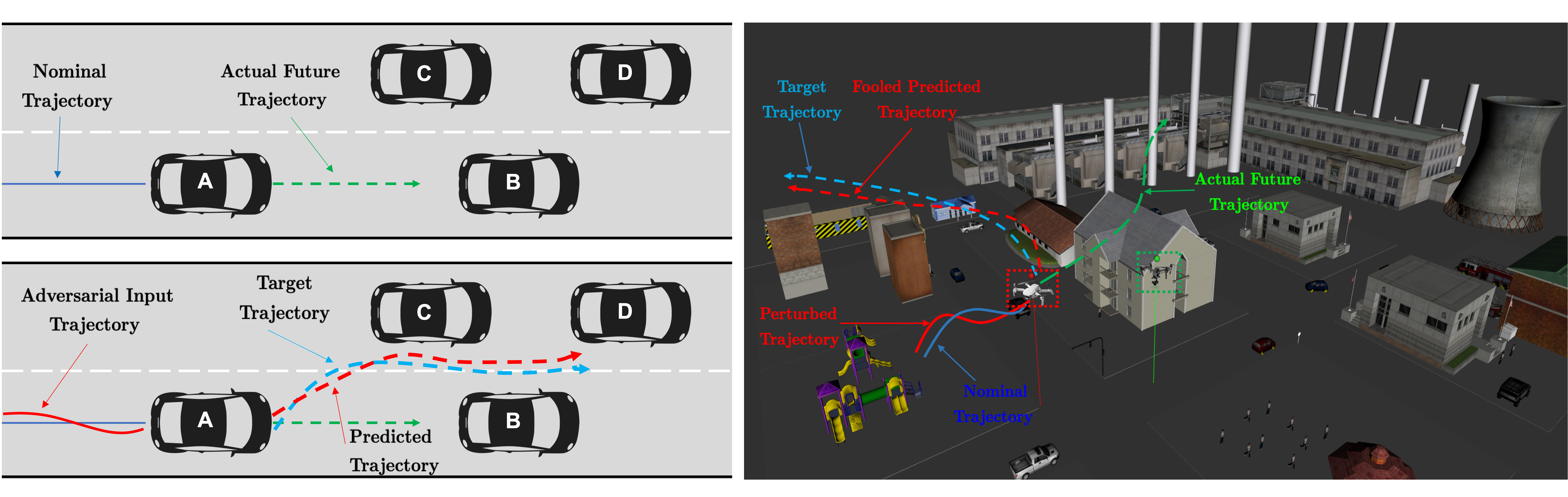}
  \caption{A graphical demonstration of the proposed \textit{targeted} adversarial attack in an autonomous driving (left) and pursuit evasion scenario (right). In the left figure, the adversary (car A) designs a trajectory so that the DNN model of car C predicts that car A will accelerate and move between cars B and C. The latter may cause the nearby moving cars make unsafe decisions (e.g., accelerate or even exit their road lanes). In the right figure, the red marked drone plans to  move towards a restricted area to take pictures of factory facilities. The green marked drone collects past trajectories of nearby moving agents and, using a DNN model, predicts their future paths. If the predicted trajectories lead towards the restricted area, an alarm is raised, and the green drone is tasked with pursuing the intruders. One of the strategies that the red drone applies to remain stealthy is to follow adversarially crafted trajectories that will make the green drone specifically predict that the red drone is heading away from the restricted area.}
  \label{fig:examples}
\end{figure}

Trajectory prediction algorithms play a pivotal role in enabling autonomous systems to make safe and efficient control decisions in highly dynamic environments as they can forecast future behaviors of nearby moving agents \cite{hewing2020simulation,peddi2020data,fridovich2020confidence,omainska2021gaussian,hosseinzadeh2021toward,zhu2021learning,schumann2022benchmark,kalluraya2022multi,lindemann2022safe,nakamura2022online,fang2022behavioral,espinoza2022deep}. 
To address the lack of knowledge of other agents' intentions and control policies, deep neural network (DNN) models are often employed to address behavior forecasting tasks 
as e.g., in \cite{nikhil2018convolutional,li2019grip++,salzmann2020trajectron++,cheng2022gatraj}. These works typically assess the performance of the proposed DNN models by measuring the deviation of the predicted trajectories from the ground truth ones. However, they neglect to evaluate their robustness against adversarially crafted input trajectories. In fact, lack of adversarial robustness can significantly compromise safety of autonomous systems; see e.g., the autonomous driving and pursuit-evasion scenarios shown in Fig. \ref{fig:examples}. 


A first step towards evaluating robustness of trajectory prediction models is to provide automated methods that compute adversarial inputs (i.e., corner cases in the input space) where these models fail. To this end, in this paper, we propose a new white box targeted adversarial attack against DNN models used for trajectory forecasting tasks. We call the proposed attack TA4TP for Targeted adversarial Attack for Trajectory Prediction. The goal of TA4TP is to perturb any nominal input trajectory so that the DNN prediction is as close as possible to a user-specified target/desired trajectory. Throughout the paper, trajectories are defined as finite sequences of system states (e.g., positions of a car). We formulate the attack design process as a constrained non-linear optimization problem where the objective function captures the deviation of the predicted trajectory from the desired one and the constraints capture physical requirements that the perturbed input should satisfy. Specifically, to define the objective function, we first assign weights to each state in the target trajectory; the higher the weight is, the more important the corresponding desired state is. This allows an adversary to assign priorities to the desired states. For instance, in certain applications it may be significant for an adversary to make other agents wrongly predict that its final state is within a certain region while the predicted trajectory towards that region may be of secondary importance; this is the case e.g., in the autonomous driving example shown in Fig. \ref{fig:examples} and in the experiments provided in Section \ref{sec:exp}. 
Then, the objective function is defined as the weighted average $\ell_2$ distance between the predicted and the desired states. The constraints require the adversarially perturbed trajectory to satisfy certain physical constraints. 
For instance, in Fig. \ref{fig:examples}, in the autonomous driving scenario, the perturbed trajectory should stay within the lane and close enough to the nominal trajectory. Similarly, in the pursuit-evasion scenario shown in Fig. \ref{fig:examples},  the perturbed trajectory should be obstacle-free. Assuming that the structure of the target DNN model is fully known, we solve this optimization problem by leveraging gradient-based methods, such as the Adam optimizer \cite{kingma2014adam}. Our experiments on state-of-the-art datasets and DNN models show that the proposed attack can successfully force given (and known to the attacker) DNN models to predict desired trajectories. We believe that the proposed attack will enable users to evaluate as well as enhance the adversarial robustness of DNN-based trajectory forecasters. 
\textbf{Related Works:} DNNs have seen renewed interest in the last decade due to the vast amount of available data and recent advances in computing. In autonomous systems, DNNs are typically used either as feedback controllers and planners \cite{gao2019reduced,bansal2020combining,pfrommer2022safe,djeumou2022learning}, perception modules \cite{redmon2016you,minaee2021image}, or for trajectory prediction \cite{cheng2022gatraj,li2019grip++,salzmann2020trajectron++} that is also the case in this paper. Despite the impressive experimental performance of DNNs, their brittleness has resulted in unreliable system behaviors and public failures 
preventing their wide adoption in safety critical applications. This is also demonstrated by several adversarial attack algorithms that have been proposed recently. These attacks, similar to the proposed one, aim to minimally manipulate inputs to DNN models, so that they can cause incorrect outputs that would benefit an adversary. The large majority of existing adversarial attacks against DNN models are focused on perceptual tasks such as image classification or object detection as e.g., in \cite{goodfellow2014explaining,carlini2017adversarial,papernot2016limitations,moosavi2016deepfool,eykholt2018robust,li2019adversarial,boloor2020attacking,choi2022argan}. Recently, adversarial attacks against DNNs used for planning and control have been proposed in \cite{huang2017adversarial,ilahi2021challenges,sarkar2022reward}. However, there is a very limited number of studies evaluating robustness of DNN models for trajectory prediction against adversarial attacks. We believe that the closest works to ours are the recent ones presented in \cite{zhang2022adversarial,cao2022advdo}.  Common in these works is that they design \textit{untargeted} attacks, i.e., they aim to maximize the prediction error or, in other words, the difference between predicted and ground truth trajectories. 
To the contrary, in this work we design \textit{targeted} adversarial attacks to make DNN  predictions be as close as possible to \textit{any} user-specified desired trajectories. 
We argue that the proposed targeted attack is more expressive than un-targeted ones as the latter do not allow the adversary to freely pick any desired predicted trajectory and, therefore, cause desired unsafe situations. For instance, using untargeted attacks, in the autonomous driving setup in Fig. \ref{fig:examples}, an adversarially crafted trajectory for car A that maximizes the prediction error may point to the left or right lane which may not necessarily compromise safety of other cars. To the contrary, the proposed attack allows the adversary to select target trajectories, as shown in Fig. \ref{fig:examples}, that may force other cars make unsafe decisions.
%
To the best of our knowledge, we propose the first \text{targeted} adversarial attack against trajectory forecasting DNN models.

\section{Problem Formulation}

In this section, we first describe the trajectory prediction task (Section \ref{sec:trajPred}) and then we formally define the targeted adversarial attack design problem as a nonlinear constrained optimization problem (Section \ref{sec:probForm}).

\subsection{Trajectory Prediction via Deep Neural Networks}\label{sec:trajPred}

\begin{figure}[H]
  \centering
    \includegraphics[width=0.9\linewidth]{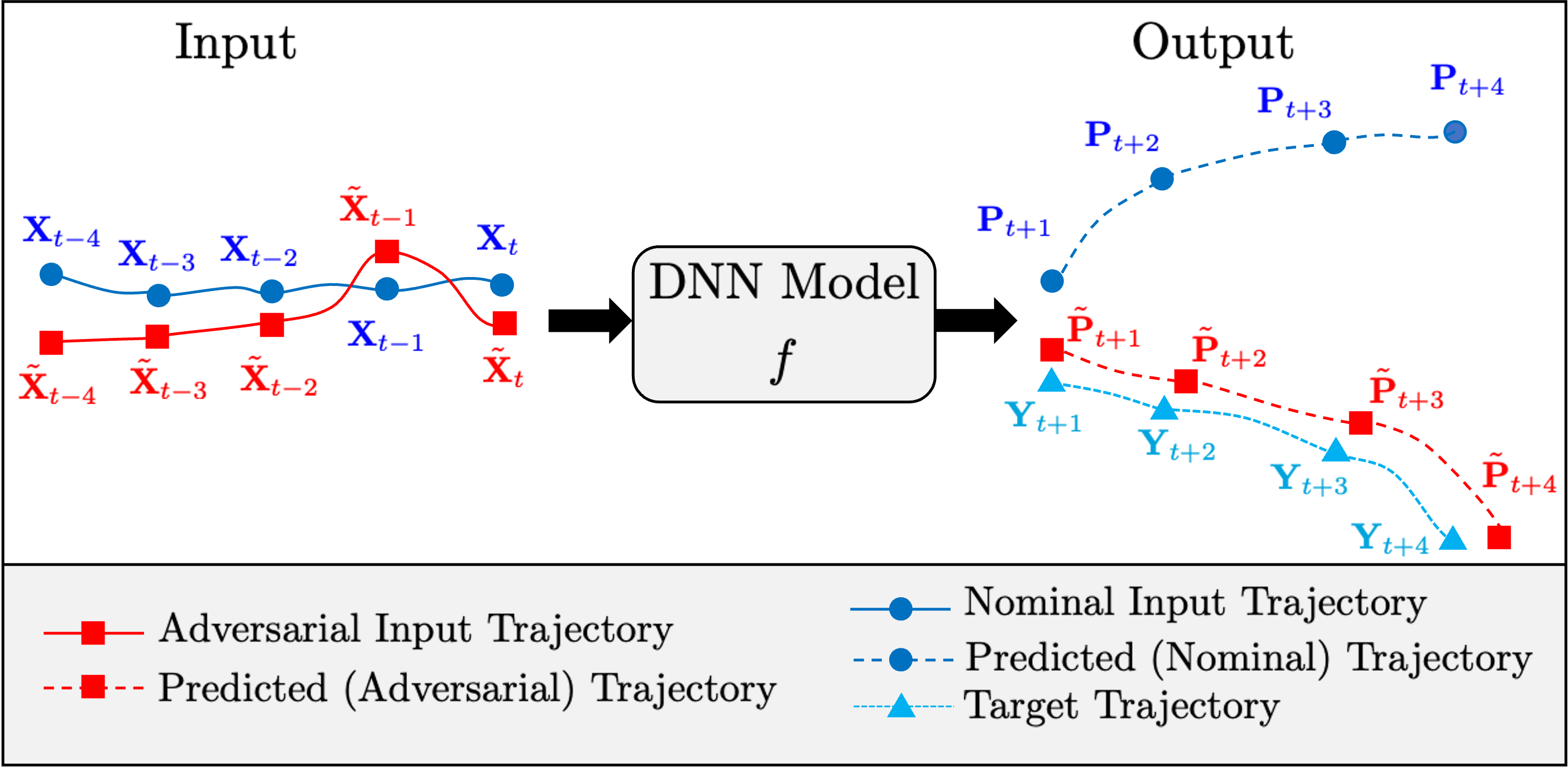}
  \caption{Graphical illustration of the problem formulation for $P=F=4$. Our goal is to design an adversarial input trajectory (red solid line) that looks natural (i.e., close to nominal inputs - blue solid line) and fools the DNN model into predicting a trajectory (red dashed line) that is as close as possible to a desired/target trajectory (cyan dashed line).}
  \label{fig:notations}
\end{figure}

We consider trajectory prediction tasks accomplished by DNNs. The goal in these tasks is to forecast the future trajectory of an agent given its past trajectories. Particularly, a DNN model takes as an input a sequence of $P$ past observed states of a moving agent (e.g., locations of a pedestrian) every $T$ time units, and outputs a sequence of predicted future states of this agent; see Fig. \ref{fig:notations}. We denote the input sequence to the DNN model by ${\bf{X}}_{t-P:t} = [{\bf{X}}_{t-P},..., {\bf{X}}_{t-1}, {\bf{X}}_t]$, where ${\bf{X}}_n$ is the state of the agent at the past time step $n\in[t-P,\dots,t]$. We also denote the ground truth future path of this agent in the next $F$ future steps as ${\bf{G}}_{t+1:t+F} = [{\bf{G}}_{t+1}, {\bf{G}}_{t+2},...,{\bf{G}}_{t+F}]$, where ${\bf{G}}_n $ stands for the ground truth state at the future time $n\in[t+1,\dots,t+F]$. Similarly, we denote the prediction of the DNN model for next $F$ steps by ${\bf{P}}_{t+1:t+F} = [{\bf{P}}_{t+1}, {\bf{P}}_{t+2},..., {\bf{P}}_{t+F}]$. Denoting the DNN model by $f$, we have that ${\bf{P}}_{t+1:t+F} =f({\bf{X}}_{t-P:t}).$

\subsection{Targeted Adversarial Attack Formulation}\label{sec:probForm}

Consider a DNN trajectory prediction model $f$ and any nominal trajectory ${\bf{X}}_{t-P:t}$ that the system has designed to follow in the time interval $[t-P,t]$. We note that ${\bf{X}}_{t-P:t}$ can be designed using any existing planning algorithm such as \cite{karaman2011sampling,janson2015fast,kantaros2020stylus}. Our goal is to design a perturbation $\boldsymbol\Delta_{t-P:t}=[\boldsymbol\Delta_{t-P}, \dots,\boldsymbol\Delta_{t-1}, \boldsymbol\Delta_{t}]$, yielding a perturbed/adversarial trajectory $\tilde{{\bf{X}}}_{t-P:t}=[\tilde{{\bf{X}}}_{t-P},\dots, \tilde{{\bf{X}}}_{t-1}, \tilde{{\bf{X}}}_{t}]$, defined as $ \tilde{{\bf{X}}}_{t-P:t}={\bf{X}}_{t-P:t}+\boldsymbol\Delta_{t-P:t}=[{\bf{X}}_{t-P}+\boldsymbol\Delta_{t-P},..., {\bf{X}}_{t-1}+\boldsymbol\Delta_{t-1}, {\bf{X}}_t+\boldsymbol\Delta_{t}]$,
so that (i) if the adversarial agent follows the trajectory $\tilde{{\bf{X}}}_{t-P:t}$, then the corresponding trajectory predicted by $f$, denoted by ${\tilde{\bf{P}}}_{t+1:t+F}=f(\tilde{{\bf{X}}}_{t-P:t})$, will be the desired/target trajectory denoted by ${\bf{Y}}_{t+1:t+F} =[{\bf{Y}}_{t+1} {\bf{Y}}_{t+2},...{\bf{Y}}_{t+F}]$ (that may be completely different from the ground truth one), i.e., ${\tilde{\bf{P}}}_{t+1:t+F}={{\bf{Y}}}_{t+1:t+F}$ 
and (ii) $\tilde{{\bf{X}}}_{t-P:t}$ satisfies certain physical constraints; see Fig. \ref{fig:notations}. To design this attack (i.e.,  $\tilde{{\bf{X}}}_{t-P:t}$), we assume that the attacker has full knowledge of the DNN model $f$. Formally, we formulate the targeted adversarial attack design problem as a nonlinear optimization problem defined as follows: 
\begin{subequations}
\label{eq:ProbR}
\begin{align}
& \min_{\substack{\boldsymbol\Delta_{t-P:t}}} J(\boldsymbol\Delta_{t-P:t})=\sum_{m=t+1}^{t+F}w_m\lVert\tilde{{\bf{P}}}_{m}-{\bf{Y}}_m\rVert_2 \label{obj}\\
& \ \ \ \ \ \ \  \tilde{{\bf{X}}}_{n}\in\mathcal{C}_n, \forall n\in[t-P,\dots,t]\label{constr}
\end{align}
\end{subequations}
{where, ${\tilde{\bf{P}}}_{t+1:t+F} =f(\tilde{{\bf{X}}}_{t-P:t})=f({\bf{X}}_{t-P:t}+\boldsymbol\Delta_{t-P:t})$. The objective function in \eqref{obj} captures the weighted average distance, using the $\ell_2$ norm, between the predicted trajectory (i.e., ${\tilde{\bf{P}}}_{t+1:t+F}$) and the desired trajectory (i.e., ${\bf{Y}}_{t+1:t+F}$). Also, in \eqref{obj}, $w_m$ is a weight modeling the importance of the $m$-th state (i.e., ${\bf{Y}}_m$) in the desired trajectory ${\bf{Y}}_{t+1:t+F}$; the higher the $w_m$ is, the more important is for $\tilde{{\bf{P}}}_{m}$ to be close to ${\bf{Y}}_{m}$. The weights are selected so that $w_m\in[0,1]$ and $\sum_{m=t+1}^{t+F}w_m=1$. The constraint $\tilde{{\bf{X}}}_{n}\in\mathcal{C}_n$, for all $n\in[t-P,\dots,t]$, requires each state $\tilde{{\bf{X}}}_{n}$ to belong to a set $\mathcal{C}_n$ collecting all permissible values. For instance, in an autonomous driving scenario, if ${\bf{X}}_{n}$ captures the agent position, then $\tilde{{\bf{X}}}_{n}\in\mathcal{C}_n$ may require the adversarially crafted trajectory to be fully within the lane (to ensure safety of the adversary). Note that, in general, the sets $\mathcal{C}_n$ can be defined differently across the states of the trajectories while their design is scenario-specific. We assume that all states in the nominal trajectory satisfy the corresponding constraints and, therefore, \eqref{eq:ProbR} is feasible; e.g., zero perturbation is a feasible solution. 
In summary, in this paper we address the following problem:}
\begin{problem}\label{pr}
\textit{Given} (i) a fully known trajectory prediction DNN model $f$; (ii) a nominal trajectory ${\bf{X}}_{t-P:t}$ that the system will follow in the time interval $[t-P:t]$; (iii) a desired predicted trajectory ${\bf{Y}}_{t+1:t+F}$; (iv) weights $w_m$ for all  $m\in[t+1,\dots,t+F]$ and a set of permissible states $\mathcal{C}_n$ for all $n\in[t-P,\dots,t]$, \textit{compute} the perturbation $\boldsymbol\Delta_{t-P:t}$ that once applied to ${\bf{X}}_{t-P:t}$ it will minimize the average weighted deviation between the DNN prediction (i.e., $\tilde{{\bf{P}}}_{t+1:t+F}=f({\bf{X}}_{t-P:t}+\boldsymbol\Delta_{t-P:t})$) and the desired trajectory (i.e., ${\bf{Y}}_{t+1:t+F}$), as captured by \eqref{eq:ProbR}.
\end{problem}
\section{Proposed Targeted Adversarial Attack for Trajectory Prediction}
In this section, we present our approach to address Problem \ref{pr}. In the rest of this section, for simplicity of notation, when it is clear from the context, we drop the dependence of trajectories on time. For instance, we simply denote the nominal trajectory by ${\bf{X}}$ instead of ${\bf{X}}_{t-P:t}$. This extends to all sequences of states and perturbation (e.g., $\boldsymbol\Delta$,  $\tilde{{\bf{X}}}$,  ${\bf{P}}$,  ${\bf{Y}}$). 

The proposed adversarial attack, called TA4TP, leverages iterative gradient-based algorithms; see Algorithm \ref{alg:attack}. We denote by $\boldsymbol\Delta^{k}$ the perturbation generated by Algorithm \ref{alg:attack} at iteration $k$. First, we randomly initialize the perturbation, denoted by $\boldsymbol\Delta^0$. Then, at every iteration $k$ of the algorithm we update $\boldsymbol\Delta^{k}$ by moving along a descent direction that minimizes the loss function $J(\boldsymbol\Delta)$. This can be achieved by simply applying a gradient descent step i.e., 
\begin{equation}\label{eq:grad}
    \boldsymbol\Delta^{k+1}=\boldsymbol\Delta^{k}-\epsilon_k\nabla J(\boldsymbol\Delta^k),
\end{equation}
where $\epsilon_k$ is a step-size. We note that any other optimization algorithm can be used to compute $\boldsymbol\Delta^{k}$ so that $J(\boldsymbol\Delta_{k+1})\leq J(\boldsymbol\Delta_{k})$, such as the Adam optimizer; see Section \ref{sec:exp}. Then, we compute the corresponding perturbed trajectory as $\tilde{\bf{X}}^{k+1}={\bf{X}}+\boldsymbol\Delta^{k+1}$.

\begin{algorithm}[t]
   \caption{TA4TP: Targeted adversarial Attack for Trajectory Prediction}
   \label{alg:attack}
\begin{algorithmic}
   \STATE {\bfseries Input:} \{Nominal trajectory $\bf{X}$, DNN $f$, Target trajectory ${\bf{Y}}$, Physical Constraints $\mathcal{C}_n$\}
   \STATE {\bfseries Output:} \{Perturbed Trajectory $\tilde{\bf{X}}$\}
   \STATE Initialize $\epsilon_0$ and $\boldsymbol\Delta^0$, and set $k=0$
   \WHILE{$(k<=K_{\text{max}}) ~\text{OR}~ (J(\boldsymbol\Delta^k)\leq \tau)$}
   \STATE Update $\boldsymbol\Delta^{k+1}=\boldsymbol\Delta^{k}-\epsilon_k\nabla J(\boldsymbol\Delta^k)$
   \IF{${\bf{X}}+\boldsymbol\Delta^{k+1}$ does not satisfy the constraints $\mathcal{C}_n$}
   \STATE Compute $\boldsymbol\theta$ as per \eqref{eq:ProbTheta} (Projection)
   \STATE Compute $\boldsymbol\Delta^{k+1}=\boldsymbol\theta \circ(\boldsymbol\Delta^{k}-\epsilon_k\nabla J(\boldsymbol\Delta^k))$
   \ENDIF
   \STATE Current perturbed trajectory $\tilde{\bf{X}}^{k+1}={\bf{X}}+\boldsymbol\Delta^{k+1}$
   \STATE $k=k+1$
   \STATE Update $\epsilon_k$
   \ENDWHILE
   \STATE Output: $\tilde{\bf{X}}=\tilde{\bf{X}}^{k+1}$
\end{algorithmic}
\end{algorithm}

Next, we check if all states in $\tilde{\bf{X}}^{k+1}$ satisfy the  constraints captured in \eqref{constr}, i.e., if  $\tilde{\bf{X}}^{k+1}_n\in\mathcal{C}_n$, for all $n$. If so, then the iteration index $k$ is updated, i.e., $k=k+1$ and we repeat the above process. Otherwise, we project $\tilde{\bf{X}}^{k+1}$ into the feasible space captured by the sets $\mathcal{C}_n$. Note that $\mathcal{C}_n$ may be high-dimensional and non-convex sets making the projection process challenging. Inspired by \cite{zhang2022adversarial}, to address this issue, we apply a simple line search algorithm. Particularly, first, we introduce parameters $\theta_n\in[0,1]$, associated with each state $\tilde{\bf{X}}_n^{k+1}$. Then, we aim to find the maximum values of $\theta_n$,  so that $\tilde{\bf{X}}_n^{k+1}={\bf{X}}_n+\theta_n\boldsymbol\Delta_n^{k+1}$ belongs to the set $\mathcal{C}_n$. In math, to perform this projection, we solve the following optimization problem:
\begin{subequations}
\label{eq:ProbTheta}
\begin{align}
& \max_{\substack{\theta_{t-P},\dots,\theta_t}} \sum_{n=t-P}^t\theta_n \label{obj4}\\
& \ \ \ \ \ \ \  {\bf{X}}_n+\theta_n\boldsymbol\Delta_n^{k+1}\in\mathcal{C}_n, \forall n\in\{t-P,\dots,t\},\label{constr2}\\
& \ \ \ \ \ \ \  \theta_n\in[0,1], \forall n\in\{t-P,\dots,t\}.\label{constr3}
\end{align}
\end{subequations}
We solve \eqref{eq:ProbTheta} by simply applying a line search algorithm. Observe that since the nominal trajectory ${\bf{X}}$ satisfies the constraint \eqref{constr},  we have that \eqref{eq:ProbTheta} is always feasible (i.e., $\theta_n=0$ is always a feasible solution). We note that the above projection process may be sub-optimal if the sets $\mathcal{C}_n$ are non-convex in the sense that there may be other points on the boundary of $\mathcal{C}_n$ that are closer to ${\bf{X}}_n+\boldsymbol\Delta_n^{k+1}$ than the ones generated by solving \eqref{eq:ProbTheta}. Once $\boldsymbol\theta=[\theta_{t-P},\dots,\theta_{t-1},\theta_t]$ is computed, we update the perturbation as 
\begin{equation}
    \boldsymbol\Delta^{k+1}=\boldsymbol\theta \circ(\boldsymbol\Delta^{k}-\epsilon_k\nabla J(\boldsymbol\Delta^k)),
\end{equation}
where $\circ$ denotes the Hadamard product (i.e., the element wise product between two vectors). Next, the iteration index $k$ is updated, i.e., $k=k+1$, and the above iteration is repeated. The algorithm terminates either after a user-specified maximum number $K_{\text{max}}$ of iterations has been reached or when the loss function $J(\boldsymbol\Delta)$ is below a desired threshold $\tau$. 

\section{Experiments}\label{sec:exp}
In this section, we evaluate the efficiency of proposed attack. In particular, in Section \ref{sec:setup}, we present the considered datasets and DNN models. In Section \ref{sec:eval1}, we evaluate the performance of the designed attack under various settings.
\subsection{Experimental Setup}\label{sec:setup}
\textbf{Models:} We consider two 
state-of-the-art and open-source trajectory prediction models. The first one is  Grip++, proposed in \cite{li2019grip++}, which achieves good performance over several datasets. Grip++ uses a graph to represent the interactions of close objects and uses an encoder-decoder long short-term memory (LSTM) model to make predictions. The second model is Trajectron++ \cite{salzmann2020trajectron++}, a modular, graph-structured model that predicts the trajectories of diverse agents while incorporating agent dynamics and heterogeneous data (e.g., semantic maps). The latter predicts multiple trajectories with probabilities and we select the trajectory with the highest probability as the final result. 

\textbf{Datasets}: In our implementation, we considered two  datasets: \text{Nuscenes} \cite{caesar2020nuscenes} and \text{Apolloscape} \cite{huang2018apolloscape}. They both collect trajectories from autonomous driving scenarios in urban areas. Particularly, \text{Nuscenes} includes past trajectories with four states (i.e., $P=4$ in Section \ref{sec:trajPred}),  future trajectories with twelve states (i.e., $F=12$ in Section \ref{sec:trajPred}), and semantic maps. \text{Apolloscape} includes past trajectories that have six states (i.e., $P=6$) and  future trajectories that also have six states (i.e., $F=6$). To provide a fair comparison across datasets and models, we neglect the semantic maps in the \text{Nuscenes} dataset.

\textbf{Physical Constraints:}  We require the adversarially crafted input trajectory $\tilde{{\bf{X}}}$ to satisfy a set of physical constraints. Specifically, recall that each state $\tilde{{\bf{X}}}_n$ in $\tilde{{\bf{X}}}$ must belong to a set $\mathcal{C}_n$. Given the autonomous driving nature of the conducted experiments, we design the sets $\mathcal{C}_n$ so that they impose constraints on the position, velocity, and acceleration (all these features are included in $\tilde{{\bf{X}}}_n$) of the adversarial vehicle, as in \cite{zhang2022adversarial}. Specifically, first we require the perturbed positions in $\tilde{{\bf{X}}}_n$ to be within $1$m from the corresponding nominal/normal positions in ${{\bf{X}}}_n$ for all $n$. Given that the urban lane width is $3.7$m and the average width of cars is about $1.7$m, this constraint requires a car not shifting to another lane if it is normally driving in the center of the lane. Additionally,  we traverse all trajectories in the testing dataset to calculate the mean $\mu$ and standard deviation $\sigma$ of (1) scalar velocity, (2) longitudinal/lateral acceleration, and (3) derivative of longitudinal/lateral acceleration. For each $\mu$ and $\sigma$, we also require the respective values of the perturbed trajectories not exceeding $\mu\pm3\sigma$. These physical constraints essentially preclude careless driving of the adversarial agent and, as a result, they have the potential to preserve stealthiness of the attack. Finally, we specify the target trajectory ${\bf{Y}}$ by determining the desired positions for the adversarial vehicle. The remaining features in the target states in ${\bf{Y}}$ (e.g., velocity and acceleration) can be computed using the desired target positions. 

 
 \begin{table}[t] 
\centering
    \resizebox{\textwidth}{!}{
        \begin{tabular}{c|c|c|c|c|c} 
                        & $\bar{J}_{\text{acc}}^{\text{nom}}$(m) & $\bar{J}_{G-Y}$(m) & $\bar{J}$ (m) [Adam] & $T$ (secs) [grad]  & $T$ (secs) [Adam]  \\\hline\hline
Grip\_apolloscape       & 0.013  & 2.363          & 0.140              & 33.342    & 19.366     \\ \hline
Grip\_nuscenes          & 0.246  & 1.527          & 0.111              & 44.173    & 12.553     \\ \hline
Trajectron\_apolloscape & 0.152  & 2.698          & 0.284              & 429.333   & 161.728    \\ \hline
Trajectron\_nuscenes    & 0.450  & 0.937          & 0.031              & 582.667   & 144.648  

        \end{tabular}
            }
\caption{Summary of results for TA4TP with $K_{\text{max}}=100$ and $\tau=0.02$m. The second, third, and fourth column show the nominal accuracy of the DNN models, the average deviation between the target and the ground truth trajectories, and the average deviation between the predicted and the target trajectories, respectively. The last two columns show the average runtime to design a single adversarial trajectory.}  
\label{tab:tab1}
\end{table}
    
\textbf{Weight Assignment}: As mentioned in \eqref{obj}, weights $w_m$ need to be assigned to each state ${\bf{Y}}_m$ in the target trajectory capturing how important is the predicted states to match with the target ones. In our setup, we define the weights so that 
$0<w_{m}<w_{m+1}$ for all $m\in\{t+1,\dots,t+F-1\}$ to give more importance to the final states in the desired trajectory. Specifically, we define the weights so that the loss function in \eqref{obj} captures the exponential moving average deviation between the predicted and the desired states. We emphasize that any other definition of weights is possible.
 

\subsection{Evaluation of TA4TP}\label{sec:eval1}
In what follows, we evaluate the performance of TA4TP on the previously described datasets and models.  
We randomly sample $100$ scenarios as test cases from each dataset. 
These trajectories are called, hereafter, test trajectories. 

\textbf{DNN Nominal Accuracy:} First, we report the performance of the DNN models in the nominal setting (i.e., without any attacks) by computing the average deviation of the predicted trajectory from the ground truth one, for every test trajectory. Specifically, we compute $ {J}_{\text{acc}}^{\text{nom}}({\bf{G}},{\bf{P}})=\sum_{m=t+1}^{t+F}\frac{{\lVert{\bf{P}}_{m}-{\bf{G}}_m\rVert_2}}{F}$
for each input test trajectory ${\bf{X}}$, where 
recall that ${\bf{P}}$ is the DNN prediction given the input ${\bf{X}}$.
Then, we compute the average of $J_{\text{acc}}^{\text{nom}}({\bf{G}},{\bf{Y}})$ across the test trajectories, denoted by $\bar{J}_{\text{acc}}^{\text{nom}}$. These results are reported in the second column of Table \ref{tab:tab1}. Note that $\bar{J}_{\text{acc}}^{\text{nom}}$ is measured in meters (m) since for its computation only the positions of the car are considered (i.e., the remaining features such as velocity and acceleration are neglected as they can be uniquely computed by the positions). The same applies to all metrics discussed in the rest of this section. 

\begin{figure}[t]
  \centering
    \includegraphics[width=0.31\linewidth]{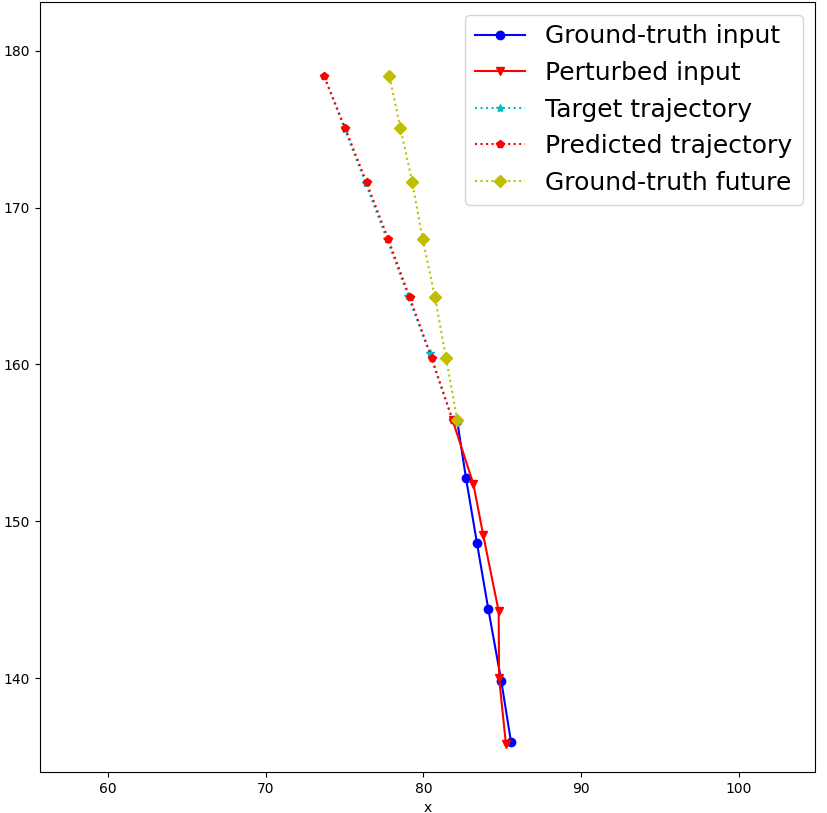}
    \includegraphics[width=0.31\linewidth]{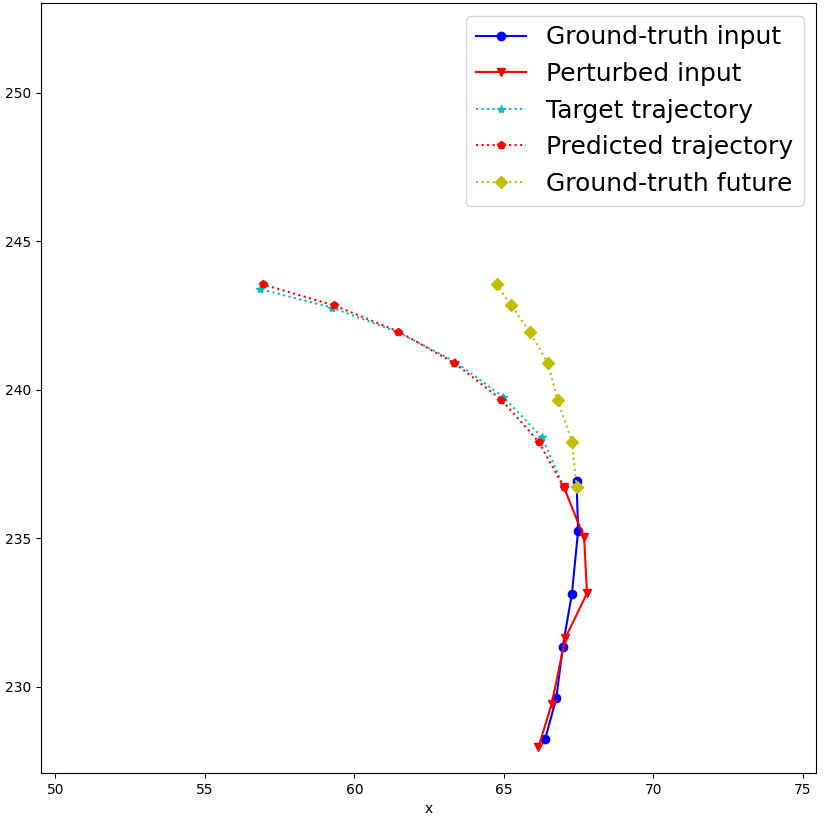}
    \includegraphics[width=0.31\linewidth]{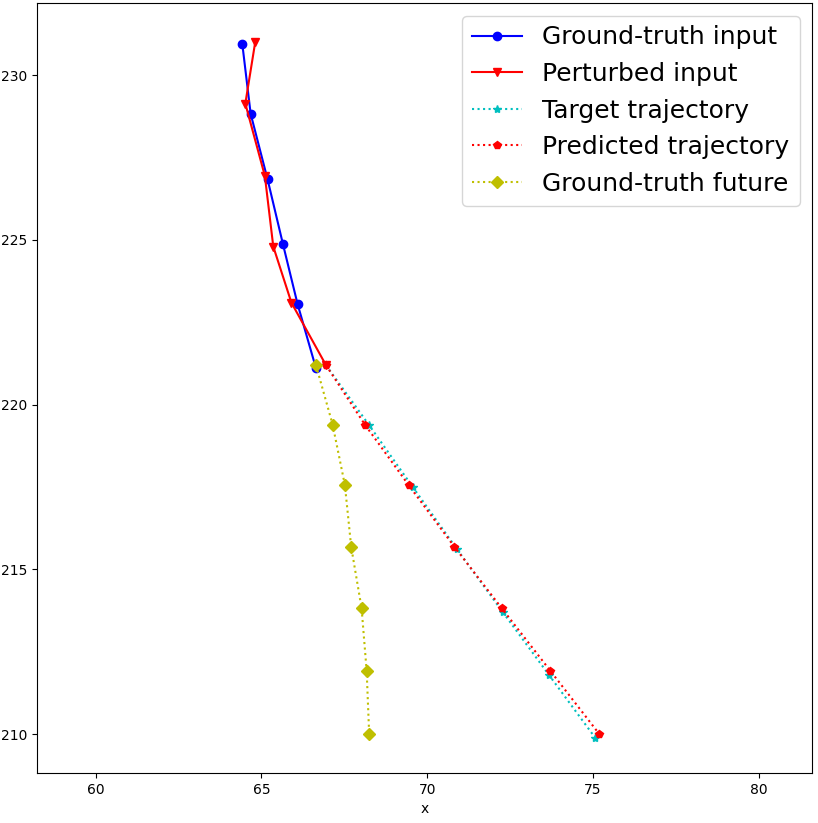}
  \caption{Graphical illustration of TA4TP in three different scenarios. Observe that in all cases the perturbed input (red solid line) is very close to the nominal trajectory (blue solid line) while the predicted trajectory (red dashed) almost overlaps with the target one (cyan dashed). }
  \label{fig:attackedl1}
\end{figure}

\textbf{Target Trajectory:} Second, we specify the target trajectories ${\bf{Y}}$. The third column in Table \ref{tab:tab1} quantifies how different the target trajectories are from the ground truth one. 
Formally, we compute the $J_{G-Y}({\bf{G}},{\bf{Y}})=\sum_{m=t+1}^{t+F}\frac{\lVert{\bf{G}}_{m}-{\bf{Y}}_m\rVert_2}{F}$, for each trajectory in the test set,
using, again, only the car positions. Then, we compute the average of $J_{G-Y}({\bf{G}},{\bf{Y}})$ across the test trajectories, denoted by $\bar{J}_{{G-Y}}$. These results are reported in the third column of Table \ref{tab:tab1}. The larger the $J_{{G-Y}}({\bf{G}},{\bf{Y}})$ is, the more the desired trajectory deviates from the ground truth.

\textbf{Evaluation of Attack Success:}
In the fourth column of Table \ref{tab:tab1}, we report the performance of TA4TP as captured by the objective function $J$ in \eqref{obj}. 
Specifically, we compute  \eqref{obj} for each test trajectory as an input and then we report the average across all trajectories, denoted by $\bar{J}$. 
The lower the $\bar{J}$ is, the more successful the attack is. 
In this setup, we terminate the optimization algorithm when either the loss function $J$ is less than $0.02$ m 
or the maximum number of iterations $K_{\text{max}}=100$ 
has been reached. Also, we used the Adam optimizer to compute $\boldsymbol\Delta^{k+1}$ (as opposed to gradient descent mentioned in Alg. \ref{alg:attack}) due to its fast convergence properties. 
Observe that good prediction accuracy on normal trajectories
does not necessarily lead to good adversarial robustness. For instance, Grip++ has a better nominal accuracy in the Apolloscape dataset than Trajectron++ does (see the second column). 

\begin{wrapfigure}{r}{7cm}
  \centering 
  \includegraphics[width=1\linewidth]{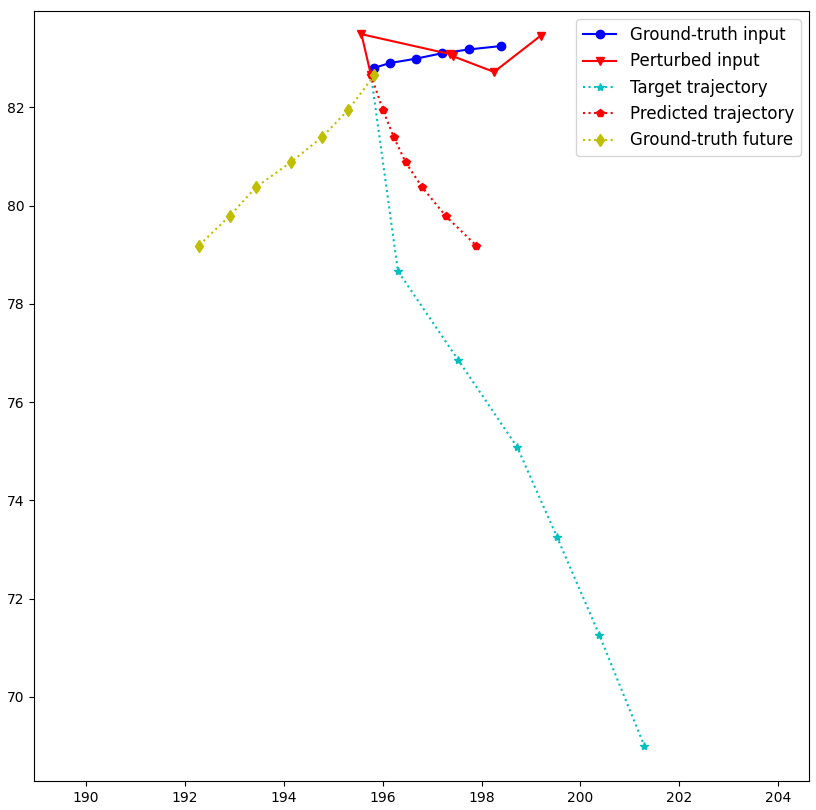}\\ 
  \caption{\footnotesize Performance of TA4TP given an aggressive target trajectory. The target trajectory requires the car to move along the cyan direction with a velocity that is significantly higher than the one associated with the input path. This difference in the velocity is illustrated by the large distance between the waypoints in the target trajectory. }\label{fig:fail} 
\end{wrapfigure}

However, performance of Grip++ on this dataset seems to be more vulnerable  than Trajectron++ to adversarial perturbations (see the fourth column). Also, observe in the fourth column that $\bar{J}$ is quite small implying that the predicted trajectories are sufficiently close to the target ones; see e.g., Figure \ref{fig:attackedl1}. 
We note that this may not always be the case depending on the physical constraints and the target trajectory. For example, if the physical constraints are very tight and the target trajectory is rather aggressive (i.e., too far from the nominal prediction), then the optimal perturbed trajectory, as per \eqref{eq:ProbR}, may not achieve a low loss as per \eqref{obj}. This is demonstrated in Fig. \ref{fig:fail}; fooling the DNN model into predicting such target trajectories requires relaxing the physical constraints.

\textbf{Attack Design Runtime:}
The last two columns in Table \ref{tab:tab1} show the average time required to generate an adversarial trajectory using gradient descent (as in \eqref{eq:grad}) and Adam optimizer. As expected, the Adam optimizer is significantly faster than the standard gradient descent method. 
Particularly, Adam reduces the computational time by at least $42\%$.
Also, notice that based on the runtimes shown in Table 1, execution of the proposed attack in real time may be prohibitive. To mitigate this, a smaller maximum number of iterations $K_{\text{max}}$ can be selected in Alg. \ref{alg:attack} which, however, may compromise the accuracy of the attack (as measured by \eqref{obj}). For instance, in Table \ref{tab:tab2}, we run the same set of experiments as before but with $K_{\text{max}}=10$. Observe that the runtimes are at least $10$ times smaller than the ones reported in Table 1 (where $K_{\text{max}}=100$). Also, observe in the second column of Table \ref{tab:tab2}, that the accuracy of TA4TP has decreased (compared to the one in Table \ref{tab:tab1}), but it still achieves a satisfactory performance. In the fourth column of Table \ref{tab:tab2}, we also report the corresponding deviation error $\bar{J}$ for the standard gradient descent approach. As expected, the Adam optimizer is faster and achieves a better performance within a fixed number of iterations. We also note that potentially an adversary can design adversarial trajectories offline, store them in a library, and  select them online when needed.

\begin{table}[t]
\centering
\resizebox{\textwidth}{!}{
\begin{tabular}{c|c|c|c|c}

                        & $\bar{J}$ (m) [Adam] & $T$ (secs) [Adam]  & $\bar{J}$ (m) [grad] & $T$ (secs) [grad]  \\ \hline\hline
Grip\_apolloscape       & 0.331       & 4.699    & 0.635      & 4.089  \\ \hline
Grip\_nuscenes          & 0.132       & 3.749    & 0.196      & 4.313  \\ \hline
Trajectron\_apolloscape & 0.364       & 35.790   & 0.495      & 35.978  \\ \hline
Trajectron\_nuscenes    & 0.077       & 39.093   & 0.199      & 46.777  
\end{tabular}
}
\caption{Summary of results for TA4TP with $K_{\text{max}}=10$ and $\tau=0.02$m.}
\label{tab:tab2}
\end{table}




 \begin{table}[b] 
\centering
    \begin{tabular}{c|c|c}

                        & $\bar{J}_{\text{acc}}^{\text{nom}}$(m) & $\bar{J}$(m) [Adam]   \\\hline\hline
Grip\_apolloscape       & 0.413     & 2.260                   \\ \hline
Grip\_nuscenes          & 1.365     & 1.219                   \\ \hline
Trajectron\_apolloscape & 0.742     & 2.153                   \\ \hline
Trajectron\_nuscenes    & 3.710     & 1.040      

        \end{tabular}   
\caption{DNN robustness analysis against random noise on clean inputs. }
\label{tab:tab3}
\end{table}
\textbf{DNN Robustness to Random Noise on Clean Inputs:} Next, we investigate how/if small random (i.e., non-adversarial) noise affects the performance of the considered DNNs in nominal  settings.  Particularly, we examine the performance of the DNN models when random noise is embedded in clean (i.e., non-adversarial) inputs. To generate a small amount of random noise, we apply the following process. Given a clean  trajectory $c$, we compute the average distance between consecutive waypoints denoted $\bar{\delta}_c$. Then for each waypoint in $c$, we sample a new waypoint within a ball centered at the original waypoint with radius $0.02\bar{\delta}_c$. These new waypoints constitute the `noisy clean' inputs that are close enough to the original ones.
Then, given these noisy inputs, we compute the average nominal accuracy $\bar{J}_{\text{acc}}^{\text{nom}}$ and the average deviation $\bar{J}$ from the target paths, as defined before; hereafter, we assume $K_{\text{max}}=100$ and $\tau=0.02$ and we consider the same target paths as the ones considered before. The results are reported in Table \ref{tab:tab3} and they should be compared against the ones in the second and fourth column of Table \ref{tab:tab1}. Observe that the nominal accuracy $\bar{J}_{\text{acc}}^{\text{nom}}$ of both models has dropped due to the random noise demonstrating their sensitivity; Grip++ seems to be more robust to random noise than Trajectron++ is.
Notice also that $\bar{J}$ is significantly high for both models meaning that simply applying random noise cannot fool them into predicting the desired trajectories. 


\textbf{DNN Robustness to Random Noise on Adversarial Inputs:} 
We repeat the same process as above but in adversarial settings. In other words, we examine how the DNN models perform in the vicinity of adversarially crafted trajectories.  Specifically, we compute $\bar{J}$ after adding a small amount of noise (exactly as discussed above) into the adversarial inputs generated for Table \ref{tab:tab1}. 
This metric for Grip++ on the Apolloscape and Nuscenes dataset is $0.203$ m and $0.253$ m, respectively. Similarly, for Trajectron++, we get that the deviation error $\bar{J}$ on the Apolloscape and Nuscenes datasets is $0.507$ m and $0.099$ m, respectively. 
This error is close to the corresponding one reported for the `noiseless' adversarial inputs on the fourth column of Table \ref{tab:tab1}. This also implies robustness of TA4TP against random noise that may occur naturally e.g., due to slippery roads or wind gusts. Additionally, by comparing these values with the respective ones for the `noisy clean' inputs (see the last column in Table \ref{tab:tab3}), we see that the DNN models seem to be more robust in the vicinity of adversarial inputs than in the vicinity of clean inputs. 
We believe that this observation may also be useful to detect adversarial inputs. Similar observations have been used to detect adversarial inputs to image classifiers \cite{meng2017magnet,kantaros2021real,nesti2021detecting,kaur2022idecode}. Specifically, to detect whether an input image is benign or not, these works investigate how the DNN output changes under transformations (e.g., compression, rotation, or adding noise) applied to the inputs.

\textbf{Effect of traffic density:} Trajectory prediction models  model the interaction among objects as a graph structure to enhance prediction performance. To study the factor of traffic density, we perform the following experiment. First we randomly sample $20$ test trajectories from the Apolloscape dataset and we compute the average deviation $\bar{J}$, defined earlier, when (i) all agents in the scene are considered versus (ii) all other agents besides the adversary and a randomly selected agent  are dropped from the scene. We denote by $\bar{J}_{\text{all}}$ and $\bar{J}_{2}$ the average deviation $\bar{J}$ in the settings (i) and (ii), respectively. As for Grip++, we get that $\bar{J}_{\text{all}}=0.215$ m and $\bar{J}_{\text{2}}=0.331$m while for Trajectron++, we get that $\bar{J}_{\text{all}}=0.034$ m and $\bar{J}_{\text{2}}=0.102$ m.
Observe that the attack remains successful in both settings in the sense that the deviation from the target trajectory is quite low. It is also worth noting that $\bar{J}_{\text{all}}<\bar{J}_{2}$, i.e., it seems to be `easier' to fool the DNN models in high traffic density scenarios. Nevertheless, this result may be specific to this experimental setup.    



\section{Conclusion}
In this paper, we proposed TA4TP, the first targeted adversarial attack for DNN models used for trajectory forecasting tasks. 
We demonstrated experimentally that TA4TP can design input trajectories that look natural and are capable of fooling DNN models into predicting desired outputs. We believe that the proposed method will allow users to evaluate as well as enhance robustness of trajectory prediction DNN models.


\bibliographystyle{IEEEtran}
\bibliography{file.bib}

\end{document}